\documentclass[lettersize,journal]{IEEEtran}
\usepackage{amsmath,amsfonts,bbm}
\usepackage{algorithmic}
\usepackage{algorithm}
\usepackage{array}
\usepackage[caption=false,font=normalsize,labelfont=sf,textfont=sf]{subfig}
\usepackage{textcomp}
\usepackage{stfloats}
\usepackage{url}
\usepackage{verbatim}
\usepackage{graphicx}
\usepackage{tabularx,longtable,multirow,caption}
\usepackage[table,xcdraw]{xcolor}
\usepackage{cite}
\usepackage{lipsum}% <- For dummy text

\begin{document}

\title{
A Multicore and Edge TPU-Accelerated Multimodal TinyML\\System for Livestock Behavior Recognition
}

\author{Qianxue~Zhang,
        Eiman~Kanjo
        % <-this % stops a space
\thanks{
Qianxue Zhang is with Medical AI Lab, Hebei Provincial Engineering Research Center for AI-Based Cancer Treatment Decision-Making, The First Hospital of Hebei Medical University, Shijiazhuang 050000, China; and the Computing Department, Imperial College London, London, UK.

Eiman Kanjo is with Professor Pervasive Sensing \& TinyML and the Head of the Smart Sensing Lab at Nottingham Trent University, Nottingham, UK (email: eiman.kanjo@ntu.ac.uk); and Provost's Visiting Professor in tinyML at Imperial College London, London, UK (email: e.kanjo@imperial.ac.uk).

© 2025 IEEE.  Personal use of this material is permitted.  Permission from IEEE must be obtained for all other uses, in any current or future media, including reprinting/republishing this material for advertising or promotional purposes, creating new collective works, for resale or redistribution to servers or lists, or reuse of any copyrighted component of this work in other works.
}% <-this % stops a space

}

\maketitle

\begin{abstract}
The advancement of technology has revolutionized the agricultural industry, transitioning it from labor-intensive farming practices to automated, AI-powered management systems. In recent years, more intelligent livestock monitoring solutions have been proposed to enhance farming efficiency and productivity. This work presents a novel approach to animal activity recognition and movement tracking, leveraging tiny machine learning (TinyML) techniques, wireless communication framework, and microcontroller platforms to develop an efficient, cost-effective livestock sensing system. It collects and fuses accelerometer data and vision inputs to build a multimodal network for three tasks: image classification, object detection, and behavior recognition. The system is deployed and evaluated on commercial microcontrollers for real-time inference using embedded applications, demonstrating up to 270$\times$ model size reduction, less than 80ms response latency, and on-par performance comparable to existing methods. The incorporation of the wireless communication technique allows for seamless data transmission between devices, benefiting use cases in remote locations with poor Internet connectivity. This work delivers a robust, scalable IoT-edge livestock monitoring solution adaptable to diverse farming needs, offering flexibility for future extensions.
\end{abstract}

\vspace{-0.03\linewidth}

\vspace{-0.07\linewidth}
\section{Introduction}
\IEEEPARstart{I}{n} the traditional livestock industry, constant human attention is essential for animal activity observation and condition assessment. In some cases, domain experts are also required to secure product safety and quality. However, the shortage of labor due to urbanization have increased operational cost, prompting farmers to explore more profitable alternatives. The rapid development of IoT devices and machine learning techniques has made it feasible for rural farms to adopt intelligent farming practices at a lower cost \cite{KUMARKASERA2024108522, patrick2024internetless}. Therefore, edge AI applicationswith sensors and cameras are integrated to automate agricultural processes through real-time decisions, enabling efficient livestock monitoring with minimal human intervention \cite{10929192, 10502951}.

Ranging from computer vision to natural language processing, modern AI algorithms have become more powerful and accurate, at the expense of increased size, computational requirements and carbon footprint. Dedicated hardware such as GPU acts as the primary choice for training complex large models, but with a price proportional to its processing capability, making it barely affordable for most farms. While cloud computing offers another possible solution, model performance might be degraded due to unreliable Internet coverage in rural areas, resulting in high latency and low throughput.
Therefore, given the financial constraints and communication restrictions in agricultural environments, relying on cloud servers or high-end hardware is challenging \cite{IDOJE2021107104}. To overcome these difficulties, TinyML has emerged as a promising solution to design an affordable and efficient system. It allows for real-time data processing and on-device inference without connecting to cloud services, thereby preserving data security. With integrated sensors and embedded systems, TinyML can achieve performance comparable to cloud and fog computing on resource-constrained hardware, while reducing latency, energy consumption and privacy risks \cite{10177729, woodward2024parallel, gibbs2024multiple, bao2025decentralised}.

To enhance livestock health, boost farming productivity, and reduce operational cost in rural farms with minimal energy consumption, we propose a low-budget, wireless-enabled, and automated sensing system for reliable animal monitoring and tracking, offering a novel IoT solution to smart livestock management. The system features three key contributions:

\vspace{-0.08\linewidth}

\begin{enumerate}
    \item TinyML model development and optimization: We designed and fine-tuned several highly efficient TinyML models tailored for complex vision and sensor-based tasks, achieving up to 270$\times$ model size reduction in model size while preserving comparable detection and classification accuracy.
    \item Multimodal behavior recognition framework: We developed a robust multimodal model that fuses image inputs and accelerometer signals to produce accurate livestock behavior recognition results under various farm conditions. By incorporating both image classification model and sensor-based behavior recognition model through late fusion technique, the system achieves improved robustness against sensor failure or occlusion.
    \item Multicore embedded edge deployment on microcontrollers: The models were deployed and evaluated on resource-constrained microcontrollers (dual-core Google Coral Dev Board Micro). The M7 core accelerated with Edge TPU (Tensor Processing Unit) enabled on-device inference with $<$80ms end-to-end latency, supporting real‑time responsiveness with minimal power consumption. The M4 core handled wireless communication to broadcast messages, eliminating the reliance on constant cloud connectivity.
\end{enumerate}

\section{Background}
\subsection{State-of-the-Art TinyML Models}
Common cloud deep learning models of today are evident with their large size, which leads to excessively long computing time and extensive hardware requirements, making relevant applications unfeasible for budget-restricted use cases such as farming. To relieve the computation burden and enable efficient algorithm execution, TinyML has been introduced to compress large models for deployment on low-power and resource-constrained platforms, such as microcontrollers and IoT devices. In recent years, various model compression techniques have been developed to improve efficiency and maintain accuracy, including pruning \cite{NIPS2015_ae0eb3ee}, quantization \cite{courbariaux2016binarized, Jacob_2018_CVPR}, neural architecture search (NAS) \cite{ren2021comprehensive}, and knowledge distillation \cite{hinton2015distilling}. These fundamental methods are widely applicable for optimizing all types of complex machine learning models to fit diverse memory constraints. However, designing a viable TinyML architecture from scratch is still computationally and time-consuming, which raises the demand for models tailored to MCUs and edge devices.

The lightweight MobileNet family \cite{howard2017mobilenets, sandler2018mobilenetv2, Howard_2019_ICCV} was designed to run efficient CNN vision models on mobile devices. Its architecture and classic building blocks were progressively adopted in different TinyML models, serving as the basis for TinyML development. Nevertheless, directly deploying them on commercial MCUs typically does not function as expected due to the strict memory limit. To address this, researchers then proposed various neural network architectures for running TinyML models on MCUs. ProxylessNAS \cite{DBLP:journals/corr/abs-1812-00332} introduced hardware-aware neural network specialization, optimizing model based on explicit hardware objectives such as latency and memory restrictions. Another prominent model is MCUNet \cite{NEURIPS2020_86c51678}, a system-algorithm co-design framework that jointly performs two-stage neural architecture search and memory-efficient inference scheduling, enabling large-scale inference on strictly power-constrained MCU. Moreover, MCUNetV2 \cite{NEURIPS2021_1371bcce} offered patch-based inference strategy to further reduce the memory footprint in TinyML models. However, prior studies mainly focused on theoretical model design while paying less attention to the actual deployment scenario. In practice, executing models on the MCU often encounters special obstacles given hardware constraints and inference library complexities \cite{fi14120363}. Although MicroNets \cite{MLSYS2021_c4d41d96} successfully deployed TinyML applications on commodity microcontrollers (STM32 MCUs), reproducing this work on other edge devices remains difficult due to compatibility issues.

% \vspace{-0.04\linewidth}
\subsection{AI Applications in Farm Sensing}
AI applications have been embraced to advance farming through sensor data accumulation, rapid data transmission and remote monitoring. The integration of sensors and cameras enables running optimized machine learning algorithms on different devices based on farming demands, thereby reducing human labor and improving the overall profit.

Wearable sensors are the main tool in animal behavior classification due to their affordable, energy-efficient, and reliable nature \cite{MAO2023108043}. The hybrid CNN-LSTM model outperformed pure CNN networks in sensor-based activity recognition, as RNNs are more effective in capturing the temporal dependencies in time series inputs \cite{LISEUNE2021106566}. Due to the scarcity and inconsistent quality of inertial sensor datasets, data augmentation is essential to improve model performance and generalization ability \cite{DBLP:journals/corr/abs-2002-12478}. Common methods include perturbation in the time and frequency domain, such as time reversal, temporal flipping \cite{9566833}, and Gaussian noise injection \cite{gao2021robusttadrobusttimeseries}.

In terms of camera-integrated algorithms, most existing studies leverage object detection models to track animals in captured images. The classic R-CNN \cite{Girshick_2015_ICCV} model used a two-stage region proposal network for classification, generating more accurate results but with a longer runtime. To improve efficiency, subsequent research focused on single-stage detectors to reduce inference time, such as SSD \cite{10.1007/978-3-319-46448-0_2} and YOLO \cite{Redmon_2016_CVPR}. Their rapidity, scalability and flexibility make them suitable for real-time applications on resource-restricted devices.

Additionally, various IoT applications for livestock monitoring have been developed to enhance farm productivity. The infrastructure involves several modules to function, including sensors or cameras to collect key parameters from the environment, microcontrollers to process inputs, and wireless communication components to transmit information \cite{9681084}. Some advanced systems exploit machine learning algorithms such as SVM and DNNs to classify animal activities and predict health outcomes, further improving the accuracy and utility of these IoT solutions \cite{9322666}. For instance, Manikanta \textit{et al.} presented a holistic cloud IoT-based animal surveillance system using sensors and Arduino UNO \cite{10498549}. However, data analytics was conducted on the cloud server rather than local device, as the models were not optimized for MCU-based inference. As a result, the effectiveness of the system hinges on cloud computing capabilities offered by the service provider, making it heavily dependent on stable internet connectivity.

% \vspace{-0.03\linewidth}
\subsection{Discussion}
Although the literature makes evident the benefits of TinyML models and farming-related AI applications, there still exist cavities to be filled. For example, the majority of applications rely on high-end hardware such as GPU or personal computers, since they can offer strong computational power to enhance system performance. Meanwhile, many IoT-based solutions tend to involve cloud services, which can be unreliable in rural areas with poor connectivity. In addition, the optimized TinyML models are usually evaluated on the cloud rather than deployed on the actual microcontrollers. Therefore, this work places extra emphasis on optimizing TinyML models and inferencing them on the MCU to evaluate their performance using embedded processors (\textit{e.g.} M4, M7 CPU and TPU). We aim to develop an integrated, MCU-based smart livestock monitoring solution that does not require constant Internet connectivity, contributing to an efficient sensing system with minimal cost.

\section{System Design}
\subsection{Structural Overview}
\begin{figure}[!htbp]
    \centering
    \includegraphics[width=1\linewidth]{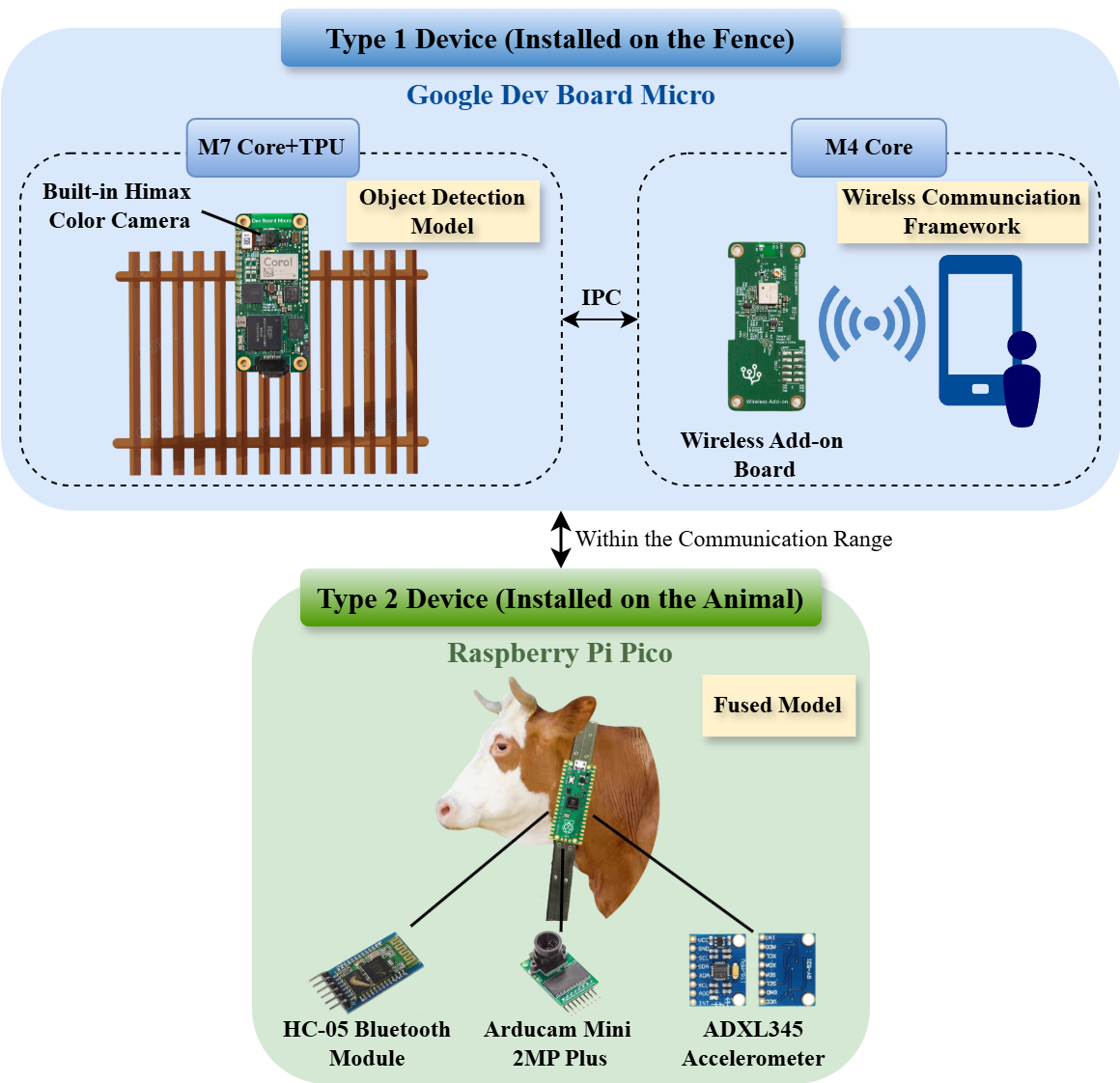}
    \caption{Structural overview of the monitoring system prototype with two types of device.}
    \label{fig:monitoring_sys}
\end{figure}

We propose a novel design for monitoring and tracking animals on the farm in Figure \ref{fig:monitoring_sys}, which consists of two types of devices that work collaboratively to track livestock behavior and health condition across a wide area. Users can leverage this system to manage a group of animals or individual ones depending on specific needs.

The static Type 1 devices are positioned on the farm's fence, integrating Google Coral Dev Board Micro to run a multicore application for real-time monitoring and communication. The M7 core employs an object detection model to track a group of animals through the built-in Himax color camera, leveraging the Edge TPU for efficient inference. Additionally, the M4 core is responsible for wireless communication with a wireless add-on module, which processes messages received from M7 and Type 2 devices to determine a notification level and transmit customized alerts to user mobile phone when abnormal behavior is detected. The two cores communicate through an inter-process communication (IPC) framework, allowing synchronized data exchange for the object detection and wireless communication tasks. 

In the meantime, the Type 2 devices (Raspberry Pi Pico) mounted on the animal's neck employ the fused model to generate detailed activity information. An Arducam Mini 2MP Plus camera and ADXL345 accelerometer are attached to collect data from the environment. By combining both vision and accelerometer inputs, the system remains robust even under restricted conditions such as partial occlusion or sensor bias. Upon identifying any concerning patterns, the device broadcasts messages to the nearby Type 1 unit to notify users through the Bluetooth module, promoting the overall efficiency and reliability. 

In terms of the hardware cost, each Type 1 device comprises a Google Coral Dev Board Micro with an Edge TPU accelerator and a wireless add-on board, at a unit price of approximately $\$100$. The Type 2 device utilizes the Raspberry Pi Pico board, combined with Bluetooth, camera, and accelerometer modules, costing around $\$25-\$30$ per unit. Therefore, for a large farm managing 100-200 animals, the overall budget would range from $\$3000-\$6000$, depending on the deployment density and communication coverage.

\vspace{-0.03\linewidth}
\subsection{Technical Details and Methodologies}
Figure \ref{fig:monitoring_sys_details} provides an in-depth illustration of the TinyML models utilized in system, outlining the expected inputs and outputs of each network. Their results are then combined and processed to generate real-time notifications through the wireless communication framework. All operations are deployed on the microcontrollers to minimize power consumption and enhance system efficiency, contributing to a reliable and cost-effective design for modern livestock management. The system comprises the following key components:

\begin{figure}[!htbp]
    \centering
    \includegraphics[width=1\linewidth]{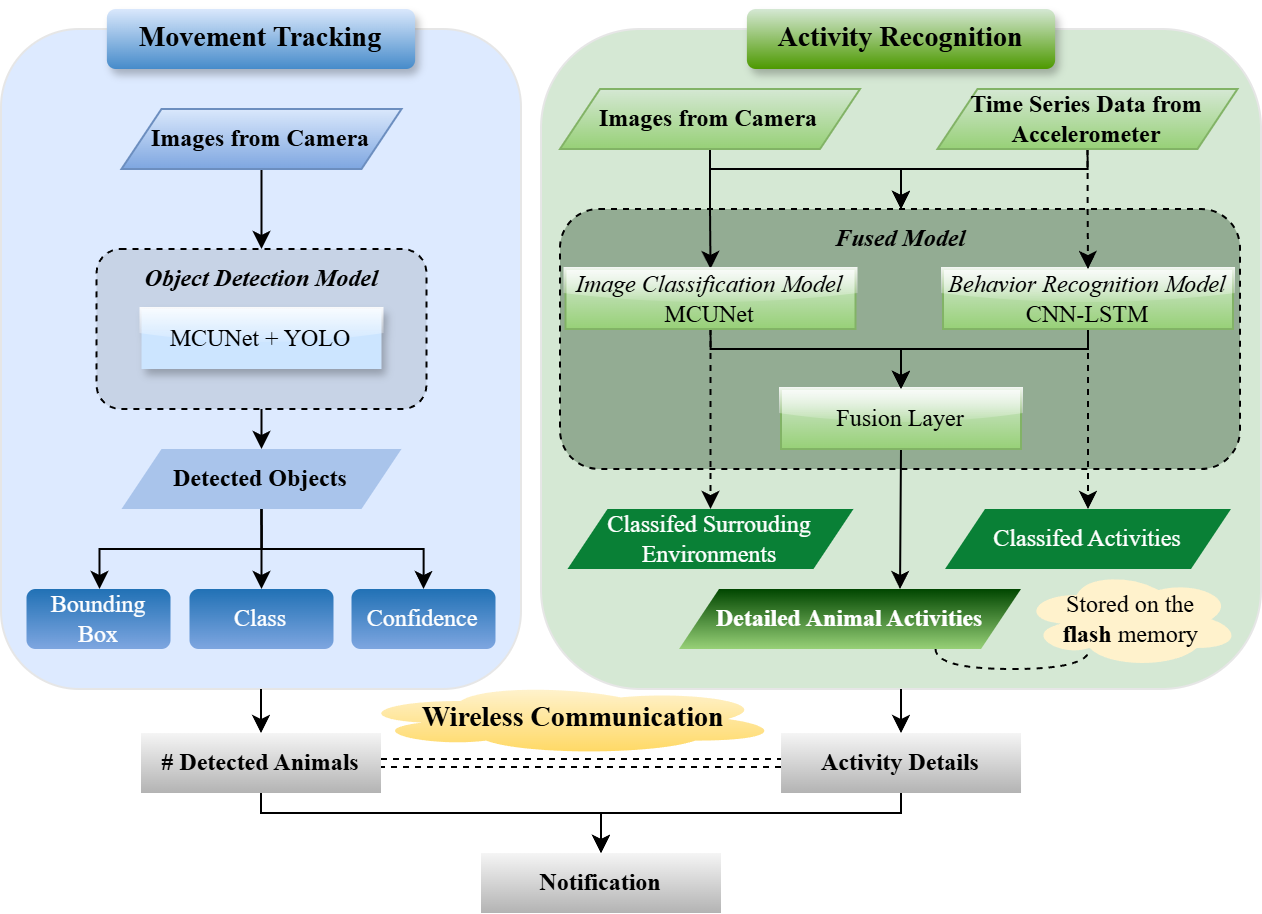}
    \caption{Detailed system architecture with movement tracking, activity recognition and wireless communication function.}
    \label{fig:monitoring_sys_details}
\end{figure}

\subsubsection{Movement Tracking}
The system employs an efficient object detection model to track animal movement, which is responsible for identifying and locating livestock within the camera frame. The MCUNet-YOLO network is optimized for low-latency inference and detection on resource-constrained devices. It outputs bounding box, class label, and confidence score for each detected object in real time, enabling continuous tracking and labeling of every animal in the frame. In addition, the detailed animal activities identified by the fused model are recorded in the device's flash memory, providing a practical solution for retrieving the livestock's actions over time and benefiting the movement tracking process. For example, the system labels a tracked animal as confined to a specific area if it feeds mostly in a stanchion without engaging in grazing activity. Conversely, if the data shows frequent grazing and moving behaviors, the animal is categorized as free-ranging.

\subsubsection{Activity Recognition}
The system integrates both sensor and camera for animal behavior recognition. The sensor-based approach serves as the primary method to classify animal activities in real-world farm environments due to its cost-effective, power-efficient and robust characteristics. A hybrid CNN-LSTM model is designed and optimized as a TinyML solution to process the time-series data collected by the accelerometer, which leverages the strengths of CNN for feature extraction and LSTM for capturing temporal dependencies in sequential data. Meanwhile, a customized image classification model, MCUNet, is employed to fully utilize vision information from the surrounding environment. Finally, the two modalities are fused at the decision level to generate more detailed and accurate activity categorizations, enhancing the overall performance and reliability of the system.

\subsubsection{Wireless Communication}
To transmit and receive messages between devices, a wireless communication framework is established by leveraging the capabilities of the existing hardware. Currently, Bluetooth Low Energy (BLE) supported by most commercial MCUs is adopted. It offers a communication range of up to 200m outdoors and is specifically optimized for low power consumption, enabling operation for months to years with a small battery.

\subsubsection{Embedded System Development}
Embedded applications are developed to deploy the TinyML models on the MCU, supporting on-device inference and other additional features such as results visualization and history storage. Since the Google Coral Dev Board Micro encompasses two CPUs (M7 \& M4), single-core and multicore designs are constructed to execute complex tasks by fully utilizing its computational capability. Google also provides the \texttt{coralmicro} APIs for FreeRTOS project development, enabling TPU acceleration and TensorFlow Lite for Microcontrollers (TFLM) usage on the MCU. Overall, the development process requires extensive C++ programming, CMake configuration for building and compilation, and TensorFlow Lite (TFLite) interpretation, integrating both hardware resources and software frameworks into a complete embedded solution.

\section{TinyML Models Development}
The essential part of the project is TinyML model exploration and optimization, aiming to train efficient and accurate models with minimal memory consumption to fit in the constrained memory of existing hardware. The primary work focuses on building and optimizing individual deep learning models for three distinct tasks.

% \vspace{-0.02\linewidth}
\subsection{Image Classification}
In the context of pasture-based farming, accurately recognizing various livestock species and their surrounding environment is critical for effective livestock management. This involves constructing a robust multi-class classification model to properly categorize images captured through on-site cameras. Additionally, depending on the location of the microcontrollers and cameras, different networks are available to provide context-aware information and enhance the system's ability. For instance, with a static camera installation on the fence, the model primarily focuses on animal identification. In contrast, when the camera is mounted on moving animals, the task shifts toward classifying the surrounding environments.

\subsubsection{Model Design}
The open-source MCUNet \cite{NEURIPS2020_86c51678} is selected as the backbone of the multi-class classification model, which is designed to enable large-scale image classification jobs on resource-constrained microcontrollers. Compared to traditional lightweight model such as MobileNetV2 or ProxylessNAS, MCUNet provides a superior trade-off between accuracy, memory footprint, and inference latency due to its hardware-aware NAS and highly optimized inference engine, making it suitable for real-time deployment on the MCU.

\begin{figure}[!htbp]
    \centering
    \includegraphics[width=1\linewidth]{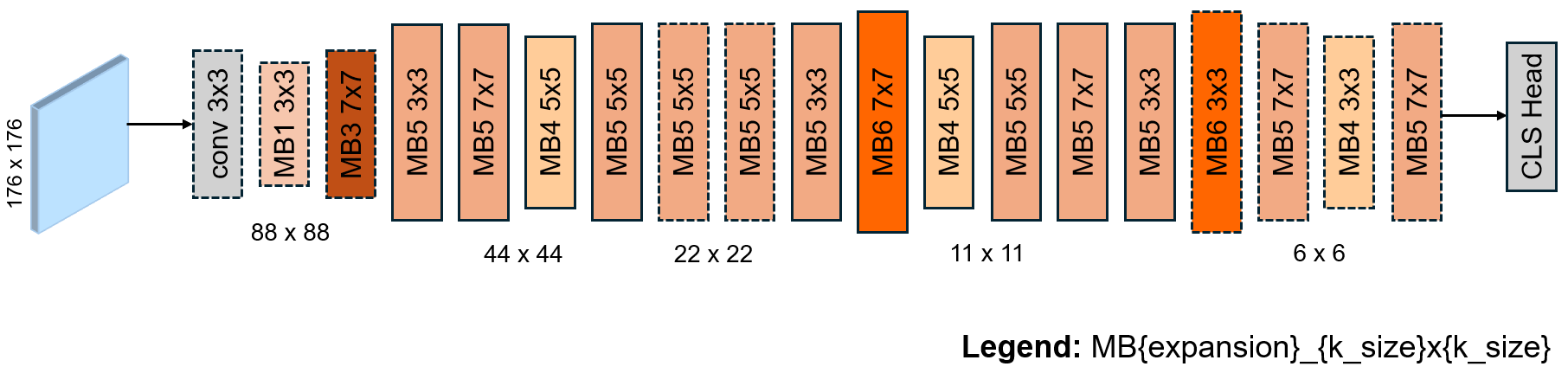}
    \caption{Architecture of the MCUNet model with MB block details.}
    \label{fig:mcunet arch}
\end{figure}

\begin{figure}[!htbp]
    \centering
    \subfloat[MobileNet core blocks.\label{fig:MB_core_block}]{
        \includegraphics[width=0.81\linewidth]{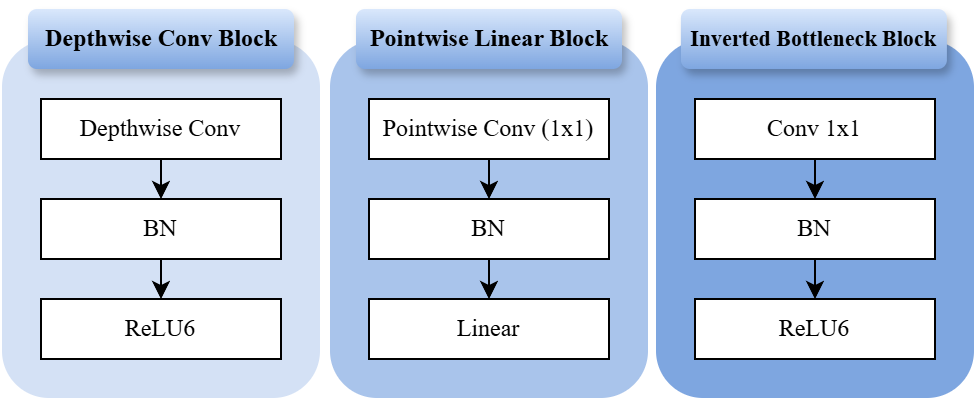}
    }
    \hfill
    \subfloat[MCUNet MB blocks.\label{fig:MCUNet_mb_blocks}]{
        \includegraphics[width=0.72\linewidth]{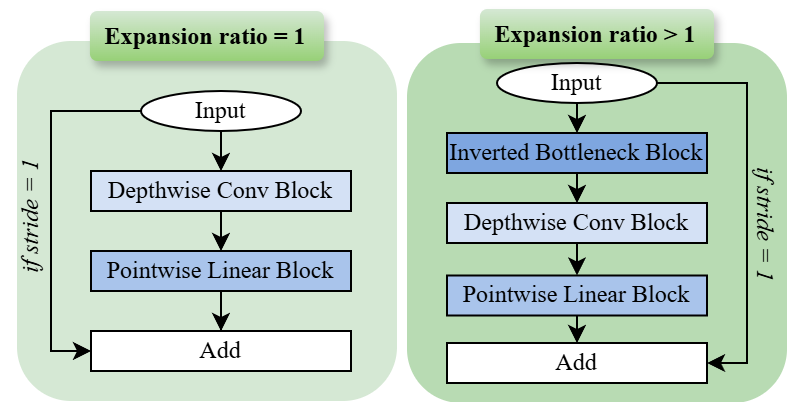}
    }
    \caption{Details of the operations in MCUNet MB blocks.}
    \label{fig:detailed_mcunet_mb_blocks}
\end{figure}

As shown in Figure \ref{fig:mcunet arch} and Figure \ref{fig:detailed_mcunet_mb_blocks}, the main architecture of the model originates from the core building blocks in MobileNetV2 \cite{sandler2018mobilenetv2}. To reduce the computational load, the input images were preprocessed to a resolution of 176$\times$176 for feature extraction. A series of expansion ratios and kernel sizes was selected in the structure based on the widely-used mobile search space \cite{DBLP:journals/corr/abs-1812-00332} and optimization results. The size of the two models is parameterized by the number of FLOPs and trainable parameters, as shown in Table \ref{img-classification-size}.

\vspace{-0.01\linewidth}

\vspace{-0.06\linewidth}

\vspace{-0.06\linewidth}
\begin{table}[!htbp]
\centering
\caption{Image classification models size.}
\label{img-classification-size}
\begin{tabular}{lll}
\hline\hline
\multicolumn{1}{c}{\textbf{Models}} & \multicolumn{1}{c}{\textbf{\#FLOPs}} & \multicolumn{1}{c}{\textbf{\#Params}} \\ \hline
MCUNet (Mini-ImageNet)              & 81.64M                               & 0.5925M                               \\
MCUNet (Custom Dataset)             & 81.64M                               & 0.5769M                               \\ \hline\hline
\end{tabular}
\end{table}

\begin{figure}[!htbp]
    \centering
    % Subfigure (a)
    \subfloat[Best $W^*$.\label{fig:best_w}]{
        \includegraphics[width=0.7\linewidth]{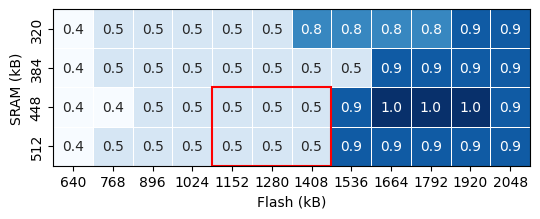}
    }
    \hfill
    % Subfigure (b)
    \subfloat[Best $R^*$.\label{fig:best_r}]{
        \includegraphics[width=0.7\linewidth]{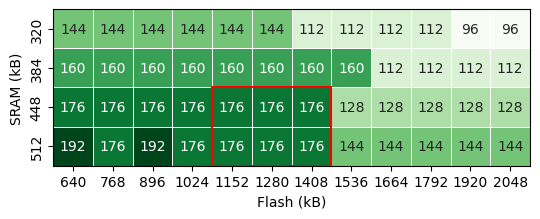}
    }
    \caption{Best search space configurations $S^*$ under various SRAM and Flash settings.}
    \label{fig:search_space_configs}
\end{figure}

% \vspace{-0.08\linewidth}
The MCUNet model was optimized by leveraging the two-stage NAS method (TinyNAS \cite{NEURIPS2020_86c51678}), which involves automated search space optimization and resource-constrained model specialization. First, based on the microcontroller resources, a broad range of width multipliers ($W$) and input resolutions ($R$) were selected as the possible search space ($S$). The best configurations $S^*$ for different SRAM and Flash combinations (Figure \ref{fig:search_space_configs}) can be efficiently retrieved by estimating the Cumulative Distribution Function (CDF) of FLOPs, since higher FLOPs generally indicate greater model capacity and accuracy \cite{he2018amc}. In order to fit into most commercial microcontrollers while considering the performance trade-off, $W^*=0.5$ and $R^*=176$ were chosen as the primary design space. Second, within $S^*$, we employed a one-shot NAS \cite{guo2020single} to train one supernet that represents all possible sub-architectures through weight sharing. The optimal model was then determined by evolutionary search to satisfy the memory constraints while maximizing accuracy.

\vspace{-0.04\linewidth}

\vspace{-0.03\linewidth}
\subsubsection{Datasets \& Training}
\begin{figure}[!htbp]
    \centering
    \subfloat[Animal.\label{fig:animal}]{
        \includegraphics[width=0.145\textwidth]{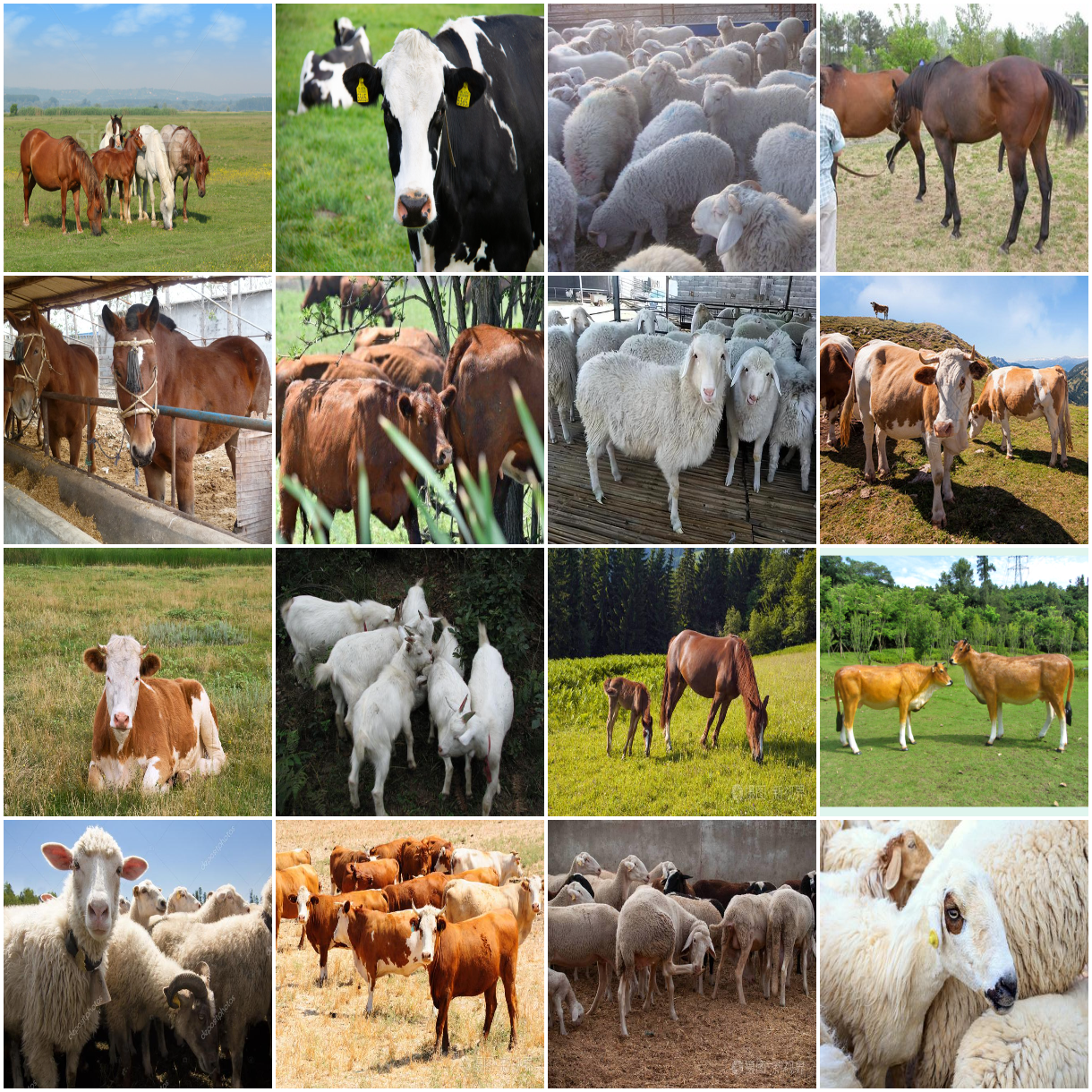}
    }
    \hfill
    \subfloat[Grass.\label{fig:grass}]{
        \includegraphics[width=0.145\textwidth]{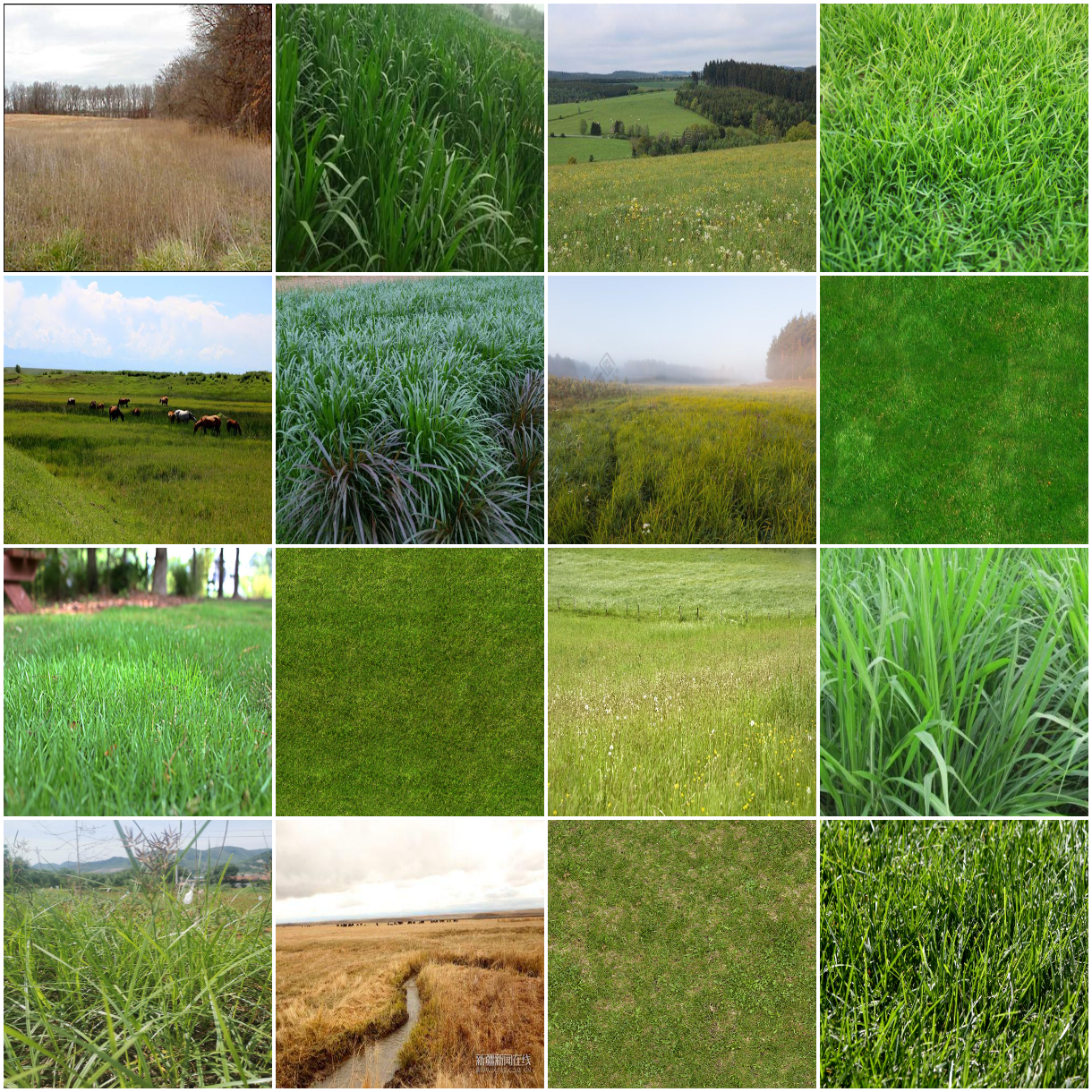}
    }
    \hfill
    \subfloat[Fence.\label{fig:fence}]{
        \includegraphics[width=0.145\textwidth]{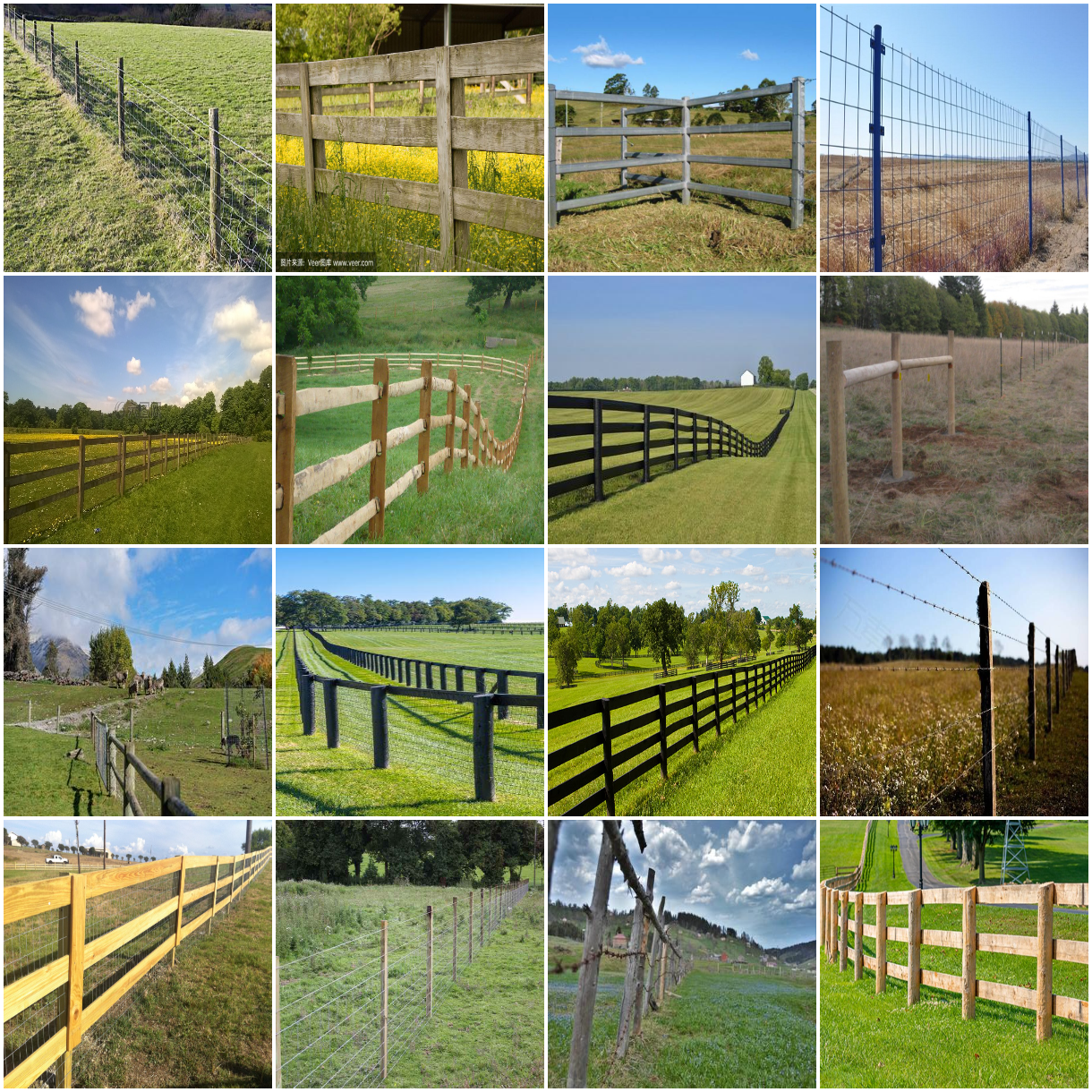}
    }
    \caption{Example images for the custom dataset.}
    \label{fig:custom_dataset}
\end{figure}

There are two datasets used for training and evaluation based on the position of the MCU. First is the mini-ImageNet dataset with fewer categories compared to the complete ImageNet dataset \cite{deng2009imagenet} , which contains 60000 images and 100 classes including common animals. The other one involves a custom dataset gathered online to identify the surrounding environment of animals, where 3 frequent scenarios are considered based on the livestock's daily routine: animal (surrounded by livestock), grass (grazing), and fence (near the farm boundary). There are 640 samples in total, which are evenly distributed among each category to ensure balance. Representative examples of each class are shown in Figure \ref{fig:custom_dataset}. To promote the robustness and generalization ability of the model, every image was carefully selected through the accumulation of varying hues, species, and photographic perspectives. 

\vspace{-0.05\linewidth}

\vspace{-0.05\linewidth}
\subsection{Object Detection}
In contrast to image classification tasks that only categorize an entire image into a single class, object detection identifies and locates multiple instances within one frame, thus being more appropriate for complex farming environments.

\subsubsection{Model Design}\label{sec:obj_det_models}
\begin{figure}[!htbp]
    \centering
    \includegraphics[width=1\linewidth]{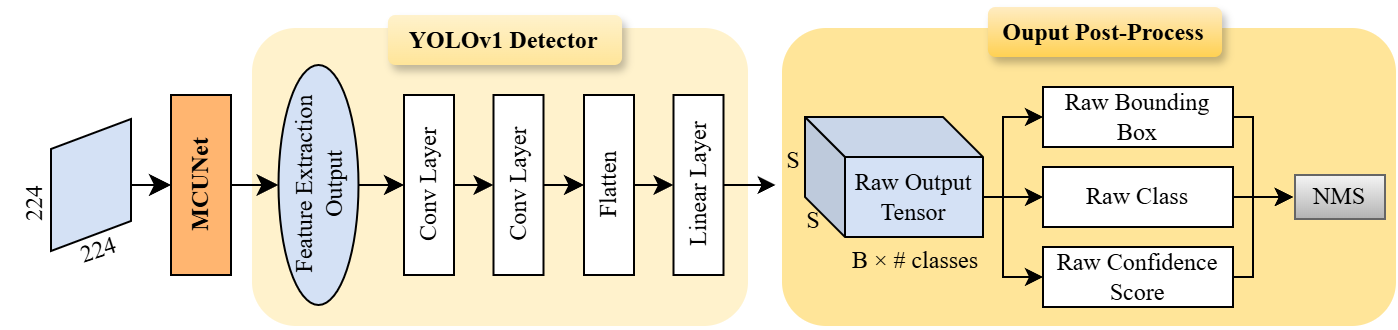}
    \caption{Architecture of the MCUNet-YOLO object detection model with YOLO detector and output post-processing steps.}
    \label{fig:yolo_mcunet}
\end{figure}

The hybrid ResNet-YOLO network in paper \cite{8833671} incorporates ResNet \cite{He_2016_CVPR} into the feature extraction part of the YOLO architecture, which is proven to be highly effective in multi-object recognition with complicated natural scenes. Inspired by this research, the MCUNet-YOLO model is designed as a TinyML solution for the object detection task. The same MCUNet continues to act as the backbone for feature extraction, and the single-stage YOLOv1 detector is employed to locate the object, resulting in a fast and memory-efficient model for real-time detection on edge devices. The input resolution was changed to 224$\times$224 based on the tuning results. Besides, the final fully connected layer was stripped off from the original MCUNet network and replaced with the corresponding YOLOv1 detector in Figure \ref{fig:yolo_mcunet}. The YOLOv1 classifier encompasses two convolutional layers and one linear layer to improve performance \cite{ren2016object}, which converts the extracted feature map from MCUNet into a raw YOLO-style tensor. The comparison of the models in terms of the number of trainable parameters and FLOPs is displayed in Table \ref{object-det-model-size}, where the MCUNet-YOLO model is much more efficient.

% \vspace{-0.04\linewidth}

\vspace{-0.04\linewidth}
\begin{table}[!htbp]
\centering
\caption{Object detection models size.}
\label{object-det-model-size}
\begin{tabular}{lll}
\hline\hline
\multicolumn{1}{c}{\textbf{Models}} & \multicolumn{1}{c}{\textbf{\#FLOPs}} & \multicolumn{1}{c}{\textbf{\#Params}} \\ \hline
ResNet-YOLO     & 16978.7M & 26.73M \\
MCUNet-YOLO     & 152.4M   & 0.73M  \\ \hline\hline
\end{tabular}
\end{table}

\vspace{-0.04\linewidth}
\subsubsection{Datasets \& Training}
The Pascal VOC dataset \cite{everingham2010pascal} includes images from a wide variety of real-world scenes. It contains 20 classes in total, covering common objects such as animals, people, and household items. The dataset is generally clear and concise, where the annotated elements are of moderate size and well-centered, occupying a significant portion of the image without complex background. The 2007 and 2012 published datasets were used to refine and evaluate the model. Since the system primarily targets the livestock use case, the number of classes was reduced to 4, consisting of cow, horse, person, and sheep only to save training time. To improve the model’s robustness and generalization ability, customized data augmentations were applied to both input images and bounding boxes to update them collectively. The pre-trained ImageNet MCUNet backbone from the image classification task was integrated into this model. 

The training loss function was implemented by following the original YOLOv1 paper \cite{Redmon_2016_CVPR}, where the network divides the image into an $S \times S$ grid, and each cell predicts $B$ candidate bounding boxes with associated confidence scores and class probabilities. The loss consists of three parts:
the bounding box regression loss that penalizes the squared error between the predicted and true bounding box center position and size, confidence loss calculated from intersection-over-union (IoU) and whether an object is present, and class prediction loss based on classification results.

\setlength{\abovedisplayskip}{5pt}
\begin{small}
\begin{multline}
L = \lambda_{\text{coord}} \sum_{i=0}^{S^2} \sum_{j=0}^{B} \mathbbm{1}_{ij}^{\text{obj}} \left[ (x_i - \hat{x}_i)^2 + (y_i - \hat{y}_i)^2 \right] \\
+ \lambda_{\text{coord}} \sum_{i=0}^{S^2} \sum_{j=0}^{B} \mathbbm{1}_{ij}^{\text{obj}} \left[ (\sqrt{w_i} - \sqrt{\hat{w}_i})^2 + (\sqrt{h_i} - \sqrt{\hat{h}_i})^2 \right] \\
+ \sum_{i=0}^{S^2} \sum_{j=0}^{B} \mathbbm{1}_{ij}^{\text{obj}} (C_i - \hat{C}_i)^2 + \lambda_{\text{noobj}} \sum_{i=0}^{S^2} \sum_{j=0}^{B} \mathbbm{1}_{ij}^{\text{noobj}} (C_i - \hat{C}_i)^2 \\
+ \sum_{i=0}^{S^2} \mathbbm{1}_{i}^{\text{obj}} \sum_{c \in \text{classes}} (p_i(c) - \hat{p}_i(c))^2
\end{multline}
\end{small}
\setlength{\belowdisplayskip}{5pt}
where:
\begin{itemize}
    \item $S \times S$ is the grid size,
    \item $B$ is the number of bounding boxes per grid cell,
    \item $(x_i, y_i, w_i, h_i)$ are the predicted bounding box coordinates and dimensions,
    \item $(\hat{x}_i, \hat{y}_i, \hat{w}_i, \hat{h}_i)$ are the ground truth bounding box parameters,
    \item $C_i$ and $\hat{C}_i$ represent the confidence scores,
    \item $\lambda_{\text{coord}}$ and $\lambda_{\text{noobj}}$ are scaling factors for localization and confidence losses, where $\lambda_{\text{coord}} = 5$ and $\lambda_{\text{noobj}} = 0.5$,
    \item $p_i(c)$ and $\hat{p}_i(c)$ are the predicted and actual class probabilities.
\end{itemize}

The input resolution can significantly affect the model performance. In the meantime, the grid size $S$ varies with the change in resolution. As summarized in Table \ref{mcunet_pascal}, a larger resolution tends to generate higher mean Average Precision (mAP) at the expense of increased computational complexity and energy consumption. Eventually, a resolution size of $224\times224$ was chosen based on the hyperparameter tuning results and computational efficiency trade-offs.

\vspace{-0.02\linewidth}
\begin{table}[ht]
\centering
\caption{MCUNet-YOLO resolution \& S tuning results.}
\label{mcunet_pascal}
\begin{tabular}{llll}
\hline\hline
\multicolumn{1}{c}{\textbf{S}} & \multicolumn{1}{c}{\textbf{Resolution}} & \multicolumn{1}{c}{\textbf{mAP}} & \multicolumn{1}{c}{\textbf{\#FLOPs}} \\ \hline
14 & 448 & 0.55 & 526.1M \\ \rowcolor[HTML]{FFFFC7}
7  & 224 & 0.49 & 152.4M \\
8  & 176 & 0.42 & 90.37M \\ \hline\hline
\end{tabular}
\end{table}

\vspace{-0.07\linewidth}
\subsection{Behavior Recognition}\label{sec:acc_classify}
Aside from vision tasks, environmental data collected from sensors is also leveraged to classify various animal activities. Accelerometer is considered the primary choice due to its affordability and ability to accurately capture livestock's behavior during movement. Besides, it is small and lightweight, which can be attached to animals with minimal impact on their natural behaviors. 

\subsubsection{Model Design}
\begin{figure}[!htbp]
    \centering
    \includegraphics[width=0.97\linewidth]{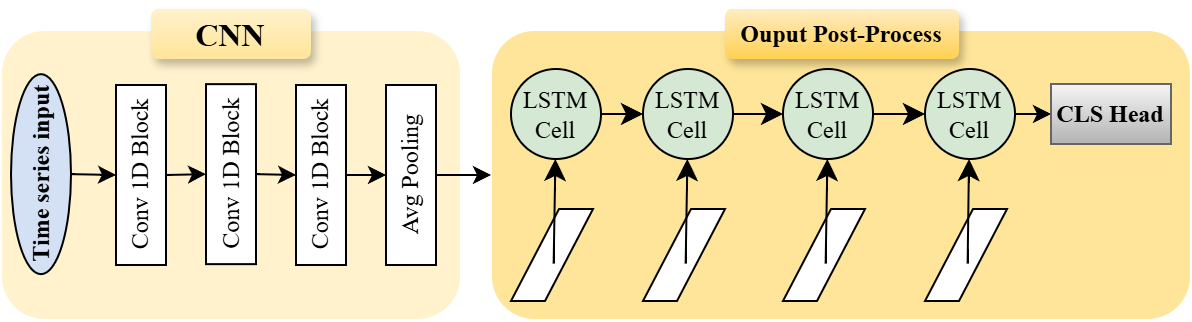}
    \caption{Architecture of the CNN-LSTM model with 3 CNN layers and 4 LSTM layers.}
    \label{fig:c_lstm}
\end{figure}

The hybrid CNN-LSTM model is incorporated due to its reliability, efficiency and high performance since it combines the strengths of both architectures to generate a stronger network \cite{MAO2023108043, LISEUNE2021106566}. It was compressed to fit into the constrained memory of edge devices, with the corresponding size in Table \ref{acc-classify-model-size}. In this architecture, CNN layers are responsible for local feature extraction, generating feature maps that represent high-level patterns through dimensionality reduction. To enhance the robustness of the model and prevent overfitting, a dropout layer is incorporated after each CNN block for regularization. LSTMs then perform sequence learning over long-term dependencies and sequential relationships in the extracted features, which is particularly useful for handling time-series data. The network structure is shown in Figure \ref{fig:c_lstm}, where the LSTM outputs is passed to the linear layer for final classification.

\vspace{-0.02\linewidth}
\begin{table}[!htbp]
\centering
\caption{Behavior classification models size.}
\label{acc-classify-model-size}
\begin{tabular}{lll}
\hline\hline
\multicolumn{1}{c}{\textbf{Models}} & \multicolumn{1}{c}{\textbf{\#FLOPs}} & \multicolumn{1}{c}{\textbf{\#Params}} \\ \hline
Original CNN              & 79.07M & 0.662M  \\
Iteratively Pruned CNN    & 25.67M & 0.1055M \\
Original CNN-LSTM           & 56.28M & 0.437M  \\
Iteratively Pruned CNN-LSTM & 12.52M & 0.1858M \\ \hline\hline
\end{tabular}
\end{table}

A grid search was performed within a range of CNN layers of $\{1,2,3,4,5\}$ and LSTM cells of $\{3,4,5,6,7\}$, and the 3-layer CNN and 4-layer LSTM configuration was selected to balance the accuracy and memory footprint. As illustrated in Table \ref{CLSTM-grid}, although more convolutional blocks are likely to generate better classification performance, the added layer contributes to less than $1\%$ accuracy gain while increasing model size by over $10\%$. Meanwhile, 3 CNN blocks provide sufficient extracted features for the LSTM module to analyze without overfitting the memory use. This model was then optimized with iterative pruning to compress its size for future deployment steps.

\vspace{-0.04\linewidth}

\vspace{-0.04\linewidth}
\begin{table}[!htbp]
\centering
\caption{CNN-LSTM grid search results with accuracy and total number of parameters.}
\label{CLSTM-grid}
\begin{tabular}{cc|ccccc}
\hline\hline
\multicolumn{1}{l}{} &
  \multicolumn{1}{l|}{} &
  \multicolumn{5}{c}{\textbf{CNN}} \\
\multicolumn{1}{l}{} &
  \multicolumn{1}{l|}{} &
  \textbf{1} &
  \textbf{2} &
  \textbf{3} &
  \textbf{4} &
  \textbf{5} \\ \hline
 &
  \textbf{3} &
  \multicolumn{1}{c|}{\begin{tabular}[c]{@{}c@{}}95.96\%\\ 0.095M\end{tabular}} &
  \multicolumn{1}{c|}{\begin{tabular}[c]{@{}c@{}}97.51\%\\ 0.120M\end{tabular}} &
  \multicolumn{1}{c|}{\begin{tabular}[c]{@{}c@{}}98.34\%\\ 0.152M\end{tabular}} &
  \multicolumn{1}{c|}{\begin{tabular}[c]{@{}c@{}}98.61\%\\ 0.185M\end{tabular}} &
  \begin{tabular}[c]{@{}c@{}}98.82\%\\ 0.218M\end{tabular} \\ \cline{3-7} 
 &
  \textbf{4} &
  \multicolumn{1}{c|}{\begin{tabular}[c]{@{}c@{}}96.01\%\\ 0.128M\end{tabular}} &
  \multicolumn{1}{c|}{\begin{tabular}[c]{@{}c@{}}97.61\%\\ 0.153M\end{tabular}} &
  \multicolumn{1}{c|}{\cellcolor[HTML]{FFFFC7}\begin{tabular}[c]{@{}c@{}}98.95\%\\ 0.186M\end{tabular}} &
  \multicolumn{1}{c|}{\begin{tabular}[c]{@{}c@{}}98.44\%\\ 0.219M\end{tabular}} &
  \begin{tabular}[c]{@{}c@{}}98.51\%\\ 0.252M\end{tabular} \\ \cline{3-7} 
 &
  \textbf{5} &
  \multicolumn{1}{c|}{\begin{tabular}[c]{@{}c@{}}95.52\%\\ 0.161M\end{tabular}} &
  \multicolumn{1}{c|}{\begin{tabular}[c]{@{}c@{}}97.29\%\\ 0.186M\end{tabular}} &
  \multicolumn{1}{c|}{\begin{tabular}[c]{@{}c@{}}97.90\%\\ 0.219M\end{tabular}} &
  \multicolumn{1}{c|}{\begin{tabular}[c]{@{}c@{}}98.40\%\\ 0.252M\end{tabular}} &
  \begin{tabular}[c]{@{}c@{}}98.17\%\\ 0.285M\end{tabular} \\ \cline{3-7} 
 &
  \textbf{6} &
  \multicolumn{1}{c|}{\begin{tabular}[c]{@{}c@{}}96.24\%\\ 0.195M\end{tabular}} &
  \multicolumn{1}{c|}{\begin{tabular}[c]{@{}c@{}}97.12\%\\ 0.219M\end{tabular}} &
  \multicolumn{1}{c|}{\begin{tabular}[c]{@{}c@{}}97.84\%\\ 0.252M\end{tabular}} &
  \multicolumn{1}{c|}{\begin{tabular}[c]{@{}c@{}}97.84\%\\ 0.285M\end{tabular}} &
  \begin{tabular}[c]{@{}c@{}}98.62\%\\ 0.318M\end{tabular} \\ \cline{3-7} 
\multirow{-8}{*}{\textbf{\begin{tabular}[c]{@{}c@{}}L\\ S\\ T\\ M\end{tabular}}} &
  \textbf{7} &
  \multicolumn{1}{c|}{\begin{tabular}[c]{@{}c@{}}95.74\%\\ 0.228M\end{tabular}} &
  \multicolumn{1}{c|}{\begin{tabular}[c]{@{}c@{}}96.85\%\\ 0.253M\end{tabular}} &
  \multicolumn{1}{c|}{\begin{tabular}[c]{@{}c@{}}97.35\%\\ 0.286M\end{tabular}} &
  \multicolumn{1}{c|}{\begin{tabular}[c]{@{}c@{}}98.17\%\\ 0.319M\end{tabular}} &
  \begin{tabular}[c]{@{}c@{}}97.23\%\\ 0.352M\end{tabular} \\ \hline\hline
\end{tabular}
\end{table}

\subsubsection{Dataset \& Training}
A publicly available the Japanese Cow behavior dataset \cite{hiroyuki_ito_2022_5849025} was leveraged for model training and evaluation. There are thirteen annotated behaviors in total, including common activities such as resting, moving, and drinking, \textit{etc}. The data are gathered using tri-axial accelerometers attached to the neck of six different cows over the course of one day, at a sampling frequency of 25Hz. There are five actions to be detected, which are resting in standing position (RES), moving (MOV), attacking (ATT), feeding in stanchion (FES), and grazing (GRZ), covering animals' daily routine and behaviors that potentially requires human attention. The raw time stamps were filtered and preprocessed into 10-s sliding windows \cite{9566833} with a step size of 25, where each window was labeled according to the majority class. To examine the effect of window length, model accuracy was evaluated across window sizes from 2 to 15 seconds. The results demonstrated that accuracy increased steadily from 2 to 10 seconds, but saturated at around 99\% for window sizes $\geq$10s. Therefore, the 10-s configuration was selected. The dataset is noticeably unbalanced, so several data augmentation strategies were exploited to perturb time positions in the training data for better generalization ability \cite{DBLP:journals/corr/abs-2002-12478}, such as randomly reversing the time series and looping the existing period to compensate for information loss. 

The training procedure is straightforward, involving hyperparameter tuning to achieve an optimal model. Aside from the standard parameters such as learning rates and optimizers, input window size is also a crucial factor that affects the model performance. A longer input sequence can preserve more long-term relationships in the data, making the network less sensitive to unexpected noise and more powerful in capturing detailed information, but at a cost of increased energy consumption and model complexity. In contrast, a smaller window size focuses on localized patterns, which reduces computation load and results in faster inference. Due to LSTM cells' ability to capture temporal dependencies, the CNN-LSTM network demonstrates relatively stable results across different window sizes. Therefore, to balance the trade-off between computational requirements and system performance, the 10-s window was selected for effective animal behavior classification.

\vspace{-0.025\linewidth}
\subsection{TFLite Model Generation and Optimization}
TensorFlow Lite (TFLite) is designed for efficient on-device inference, optimized for low-power edge devices and MCUs, whose cross-platform compatibility enables deployment across various hardware architectures. To generate a TFLite model while preserving performance, the original PyTorch model was converted to TensorFlow format, followed by transformation into TFLite with specific optimizations. 

First, we manually re-implemented the complex and highly customized PyTorch models (\textit{e.g.} MCUNet) in TensorFlow format. The entire network was rebuilt properly with Keras or TensorFlow operations, while ensuring proper weight transfer through tensor permutation to address framework discrepancies. The \texttt{tf.function} decorator was further employed to generate static computation graphs, enabling framework-specific optimizations based on hardware requirements. Although this involves extensive programming work, it reduces the total number of operations and allows the flexibility of adjusting every aspect of the model to suit the target platform compared to straightforward approaches such as ONNX (Open Neural Network Exchange).

Additionally, the TFLite model was generated using the \texttt{TFLiteConverter}, incorporating optimizations to reduce model size and inference latency. We employed quantization to reduce the precision of parameters from 32-bit floating-point to 8-bit integers, making the model more suitable for memory-constrained MCUs. This is achieved using the equation:

\setlength{\abovedisplayskip}{1pt}
\begin{equation}
Q_x = \text{round} \left( \frac{x}{S} \right) + Z
\end{equation}
\setlength{\belowdisplayskip}{1pt}
where:
\begin{itemize}
    \item $Q_x$ is the quantized integer value,
    \item $x$ is the original floating-point value,
    \item $S$ is the scale factor,
    \item $Z$ is the zero-point offset.
\end{itemize}

This transformation guarantees efficient storage and computation while maintaining accuracy. A representative dataset was involved to calibrate activations during quantization, ensuring minimal loss in performance. The full integer quantization process significantly enhances model efficiency, particularly on hardware optimized for integer tensor operations, such as TPUs. Moreover, to fully extend the model's functionality, we developed custom operators in TensorFlow, then implemented and registered the corresponding C++ kernels via \texttt{OpResolver} for proper TFLite interpretation.

\section{System Implementation}
The system can be assembled after the completion of the individual TinyML models, comprising three main aspects: a multimodal network to combine both sensor (accelerometer) and camera inputs for more accurate classification, the embedded system development to support on-board inference with additional functions, and a wireless communication module to facilitate information transmission between devices. 

\vspace{-0.04\linewidth}
\subsection{Multimodal Fusion}\label{sec:multi-modal}
Based on the system design outlined in Figure \ref{fig:monitoring_sys_details}, a fused model is employed in the Type 2 devices to support the animal monitoring process with different modalities. It leverages time-series data from the embedded sensor and images captured from the real-time camera to interpret more accurate classification results tailored to agricultural demands. In this system, two TinyML models are combined using the decision-level fusion to build a multimodal network for farm management, which consists of:

\begin{itemize}
    \item An image classification model pre-trained on the custom dataset for surrounding environment categorization.
    \item A behavior recognition model based on accelerometer inputs for livestock activity recognition.
\end{itemize}

This late fusion approach is selected due to its modularity, flexibility, and robustness, achieving 97.6\% accuracy with 0.68M parameters. Each modality can be developed and optimized independently, making it easier to incorporate new modalities or update existing ones even if conflicts occur. In contrast, early fusion combines data at the feature level, where features from each modality are extracted and integrated before being fed into a single prediction model. Despite the closer relationships between modalities with early fusion approach, the model size is increased by the integration of an extra model. 
In addition, intermediate fusion aligns modality features using a shared Transformer model and integrates them through attention mechanisms. Empirical results indicated that at least four linear layers were required, leading to approximately 13\% parameter increase but less than 1\% accuracy improvement. Since the internal features must be jointly processed, the previously converted TFLite models for image classification and behavior recognition cannot be reused, making deployment on the MCU more challenging. Similarly, hybrid fusion that combines the above techniques attained similar accuracy of around 98\% but with much higher complexity. In contrast, latter fusion solution can achieve comparable performance while remaining a compact and extendable architecture, making it more suitable for resource-constrained environments.

\begin{figure}[!htbp]
    \centering
    \includegraphics[width=0.8\linewidth]{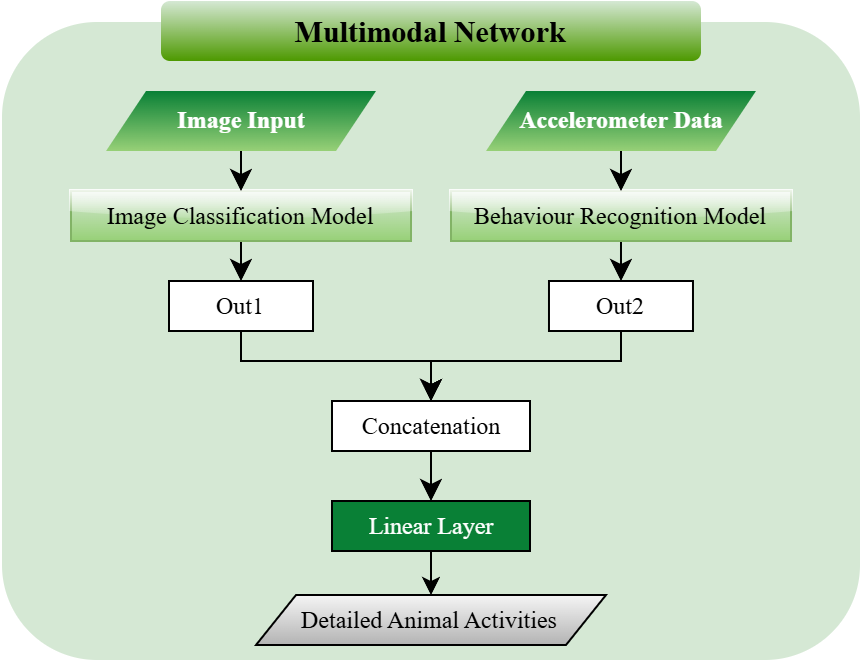}
    \caption{Architecture of the multimodal network with late fusion strategy.}
    \label{fig:fused_model}
\end{figure}

Figure \ref{fig:fused_model} illustrates the structure of the multimodal system, where each classification model is analyzed separately with its own specialized algorithms. The individual predictions are then concatenated and fed into a small linear layer to produce a final decision, represented by formula (\ref{fusion-formula}). The fusion weights were learned through a linear layer during training based on labeled multimodal data, which allows the system to automatically determine the importance of each modality in the final prediction. Finally, the fused model is converted to TFLite format with INT8 quantization for on-board deployment, ensuring efficient storage and computation on memory-constrained microcontrollers.

\setlength{\abovedisplayskip}{1pt}
\begin{equation}\label{fusion-formula}
F = \sum_{i=1}^{n} w_i \cdot P_i
\end{equation}
\setlength{\belowdisplayskip}{1pt}
where:
\begin{itemize}
    \item $F$ is the final fused decision score,
    \item $w_i$ is the weight assigned to each model's prediction,
    \item $P_i$ is the probability output of the $i$-th model.
\end{itemize}

\vspace{-0.04\linewidth}

\vspace{-0.02\linewidth}
\subsubsection{Label Mapping}
As shown in Table \ref{tab:fused-label}, the mapping of the fused labels is generated based on the classification outcomes of the two models. The first row (RES, MOV, ATT, FES, GRZ) represents the five possible livestock behaviors, while the first column (Animal, Grass, Fence) indicates the detected surrounding scenes. There are eight classes in Table \ref{tab:fused-label-meaning}, corresponding to the detailed activity information and the animal's conditions. These classes describe not only the livestock’s actions but also the level of human attention required, which is color-coded into three categories. Green denotes a safe state without immediate intervention, yellow signifies that the animal is near the fence that may need closer monitoring, and red highlights abnormal or concerning behaviors that require prompt action. For instance, the animal is considered safe when it is surrounded by animals (Animal) and resting (RES), which is labeled with 0. However, if aggressive actions are identified (ATT), or animals are moving (MOV) toward the farm's border (Fence), the manager should be notified for closer supervision. When yellow or red behaviors are detected, the system sends a message to the Type 1 device for further processing.

\vspace{-0.02\linewidth}
\begin{table}[!htbp]
\centering
\caption{Label mapping based on the animal behavior and its surrounding environment.}
\label{tab:fused-label}
\begin{tabular}{|l|l|l|l|l|l|}
\hline
                   & \textbf{RES(0)} & \textbf{MOV(1)} & \textbf{ATT(2)} & \textbf{FES(3)} & \textbf{GRZ(4)} \\ \hline
\textbf{Animal(0)} & 0               & 1               & 6               & 2               & 3               \\ \hline
\textbf{Grass(1)}  & 0               & 1               & 6               & 2               & 3               \\ \hline
\textbf{Fence(2)}  & 4               & 7               & 7               & 2               & 5               \\ \hline
\end{tabular}
\end{table}

\vspace{-0.04\linewidth}
\begin{table}[!htbp]
\centering
\caption{Detailed animal activities with labels and definitions.}
\label{tab:fused-label-meaning}
\begin{tabular}{|c|l|}
\hline
\textbf{Label} & \multicolumn{1}{c|}{\textbf{Class Meaning}} \\ \hline
\rowcolor[HTML]{B9E0A5} 
\textbf{0} & resting in safe condition                   \\ \hline
\rowcolor[HTML]{B9E0A5} 
\textbf{1} & moving in safe condition                    \\ \hline
\rowcolor[HTML]{B9E0A5} 
\textbf{2} & feeding in stanchion in safe condition      \\ \hline
\rowcolor[HTML]{B9E0A5} 
\textbf{3} & grazing in safe condition                   \\ \hline
\rowcolor[HTML]{FFF4C3} 
\textbf{4} & resting near fence                          \\ \hline
\rowcolor[HTML]{FFF4C3} 
\textbf{5} & grazing near fence                          \\ \hline
\rowcolor[HTML]{F8CECC} 
\textbf{6} & attacking animals                           \\ \hline
\rowcolor[HTML]{F8CECC} 
\textbf{7} & potential escaping                          \\ \hline
\end{tabular}
\end{table}

\vspace{-0.06\linewidth}
\subsection{Embedded System Development}
This section demonstrates the firmware development to flash the TFLite models on the MCU for inference, especially targeting the Google Coral Dev Board Micro. As the primary processor is a microcontroller, it is designed to execute tiny applications without a general-purpose operating system like Linux. Instead, all platforms and APIs are built with FreeRTOS in C++. By using the TFLM framework, on-board inference as well as other additional functions can be implemented to realize the livestock monitoring system on resource-constrained devices.

\subsubsection{Single-Core Applications}
To execute a TinyML model on the MCU, single-core applications need to be constructed based on the running processor. Since the Google Coral Dev Board Micro possesses two MCU cores, M4 and M7 with the Edge TPU accelerator, each is tested separately to measure performance. The general procedure for initiating the TFLite model inference on the Google Coral Dev Board Micro is defined in Algorithm \ref{alg:single-core} by leveraging the TensorFlow interpreter provided in TFLM. Eventually, the fused model is deployed using the single-core algorithm with TPU acceleration.

% \vspace{-0.02\linewidth}
\begin{algorithm}[!htbp]
\caption{General Single-Core Algorithm}\label{alg:single-core}
\begin{algorithmic}
\footnotesize
\STATE \textbf{Initialization:}
\STATE \hspace{0.5cm} // Power on the Edge TPU or M4 core
\STATE \hspace{0.5cm} TPU: $EdgeTpuManager \rightarrow OpenDevice()$
\STATE \hspace{0.5cm} M4: $IpcM7 \rightarrow StartM4()$
\STATE \hspace{0.5cm} // Define the TF ops
\STATE \hspace{0.5cm} $resolver.AddCustom(kCustomOp, RegisterCustomOp())$
\STATE \hspace{0.5cm} $resolver.AddConv2D()$, $resolver.AddFullyConnected()$, ...
\STATE \hspace{0.5cm} // Load TFLite model
\STATE \hspace{0.5cm} $coral\_micro::LfsReadFile(fused_model)$

\STATE
\STATE \textbf{Main Loop:}
\STATE \hspace{0.5cm} \textbf{while} \textsc{True} \textbf{do}
\STATE \hspace{1.0cm} // Calculate the inference time
\STATE \hspace{1.0cm} $dtime$ $\gets$ $t\_current$ - $t\_prev$
\STATE \hspace{1.0cm} // Prepare input tensor
\STATE \hspace{1.0cm} $input\_tensor$ $\gets$ $quantized\_input$

\STATE \hspace{1.0cm} // Invoke inference
\STATE \hspace{1.0cm} $interpreter.Invoke()$

\STATE \hspace{1.0cm} // Output post-processing
\STATE \hspace{1.0cm} $output\_tensor$ $\gets$ $dequantized\_output$
\STATE \hspace{0.5cm} \textbf{end while}
\end{algorithmic}
\end{algorithm}

% \vspace{-0.05\linewidth}
\subsubsection{Multicore Applications}
The Google Coral Dev Board Micro encompasses a dual-core MCU with an M7 and an M4 processor, providing the opportunity to run a variety of multicore applications. To fulfill the functionality of the system, we run different algorithms on the two processors and leverage the IPC feature to facilitate data sharing between them. Due to the limited computational power of the M4 core, the larger model is executed on the M7 processor, where it can use the Edge TPU for enhanced system efficiency.

The M7 core runs an object detection model (MCUNet-YOLO) accelerated by the Edge TPU to locate animals in real-time camera frames, displaying the instant results on a web interface via the local host and sending the information to the M4 core for further processing steps. The program details are shown in Algorithm \ref{alg:multi-m7}.

\vspace{-0.01\linewidth}
\begin{algorithm}[H]
\caption{M7 Algorithm (Object Detection with TPU)}\label{alg:multi-m7}
\begin{algorithmic}
\footnotesize
\STATE \textbf{Initialization:}
\STATE \hspace{0.5cm} // Set up HTTP server with URI handler
\STATE \hspace{0.5cm} $http\_server.AddUriHandler(UriHandler)$
\STATE \hspace{0.5cm} // Create synchronization primitives
\STATE \hspace{0.5cm} $img\_mutex \leftarrow xSemaphoreCreateMutex()$
\STATE \hspace{0.5cm} $bbox\_mutex \leftarrow xSemaphoreCreateMutex()$
\STATE \hspace{0.5cm} // Initialize camera in streaming mode
\STATE \hspace{0.5cm} $CameraTask \rightarrow Enable(CameraMode::kStreaming)$
\STATE \hspace{0.5cm} // Start coprocessor
\STATE \hspace{0.5cm} $IpcM7 \rightarrow StartM4()$

\STATE
\STATE \textbf{Main Loop:}
\STATE \hspace{0.5cm} \textbf{while} \textsc{True} \textbf{do}
\STATE \hspace{1.0cm} // Capture frame
\STATE \hspace{1.0cm} $CameraTask \rightarrow GetFrame(fmt)$
\STATE \hspace{1.0cm} // Run inference
\STATE \hspace{1.0cm} $interpreter.Invoke()$
\STATE \hspace{1.0cm} // Process results
\STATE \hspace{1.0cm} \textbf{return} $id, score, x\_min, y\_min, x\_max, y\_max$
\STATE \hspace{1.0cm} // IPC communication
\STATE \hspace{1.0cm} $ipc\rightarrow SendMessage(msg)$
\STATE \hspace{0.5cm} \textbf{end while}
\end{algorithmic}
\end{algorithm}

When it receives activity information from the Type 2 device and detects any animal with the M7 core, a notification level $N$ is calculated in M4 following the equation below. The higher number represents a more serious situation, where 0 indicates there is no alert to send but 3 requires immediate human intervention.
\setlength{\abovedisplayskip}{1pt}
\begin{equation}
N = \max(3, \lfloor \frac{A - 4}{2} \rfloor + \lfloor \frac{B}{2} \rfloor)
\end{equation}
\setlength{\belowdisplayskip}{1pt}

where:
\begin{itemize}
    \item $A$ is the severity score from the Type 2 device (animal activity analysis in Table \ref{tab:fused-label-meaning}),
    \item $B$ is the number of detected animals from the M7 core (object detection output).
\end{itemize}

If $N\geq1$, a BLE Beacon is set up to broadcast the customized message towards nearby mobile devices to notify users, as shown in Algorithm \ref{alg:multi-m4}. 

\vspace{-0.02\linewidth}
\begin{algorithm}[H]
\caption{M4 Algorithm (Handle msg from M7 and Type 2 Device and broadcast)}\label{alg:multi-m4}
\begin{algorithmic}
\footnotesize
\STATE \textbf{Message Handling:}
\STATE \hspace{0.5cm} // Receive M7 message and send ACK back
\STATE \hspace{0.5cm} $M7\_msg = reinterpret\_cast<const\ Message^*>(data)$
\STATE \hspace{0.5cm} $IpcM4\rightarrow SendMessage(ack\_msg)$

\STATE
\STATE \textbf{Data Processing:}
\STATE \hspace{0.5cm} // Process for notification with Type 2 Device inputs
\STATE \hspace{0.5cm} $processed\_msg \leftarrow M7\_msg, activity\_msg$
\STATE \hspace{0.5cm} $notification \leftarrow processed\_msg$

\STATE
\STATE \textbf{Bluetooth Broadcast:}
\STATE \hspace{0.5cm} // Set up a Bluetooth Beacon for advertising information
\STATE \hspace{0.5cm} $InitEdgefastBluetooth(bt\_ready)$
\STATE \hspace{0.5cm} $BluetoothAdvertise(bt\_le\_ext\_adv\ *notification)$
\end{algorithmic}
\end{algorithm}

\vspace{-0.04\linewidth}
\subsubsection{Small On-board Database}
The Coral board boasts more than 1GB of flash memory and supports local file storage through Filesystem APIs. Leveraging this capability, a lightweight on-board database is developed to store recent activity history of the monitored livestock. The database occupies $<$10KB of flash memory to store compressed activity summaries, with new messages added every 5 minutes and up to 100 entries retained using a FIFO strategy. Additionally, since the memory bottleneck on MCU typically arises from the peak activation size in SRAM rather than the flash storage \cite{lin2023tiny}, the presence of this database should not compromise the runtime performance.

This capability allows farmers and customers to access verifiable information about the animals' product history, such as whether they were free-ranging or restricted to a certain area, encouraging more informed and proactive product control. This approach provides an efficient offline method of monitoring and logging data on resource-constrained devices, significantly reducing operational latency while enhancing privacy and security through controlled access. In contrast to cloud-based solutions, the data always remains on the device, eliminating the risk of interception or unauthorized access during network transmission.

\vspace{-0.04\linewidth}
\subsection{System Construction}
\begin{figure}[!htbp]
    \centering
    \includegraphics[width=0.95\linewidth]{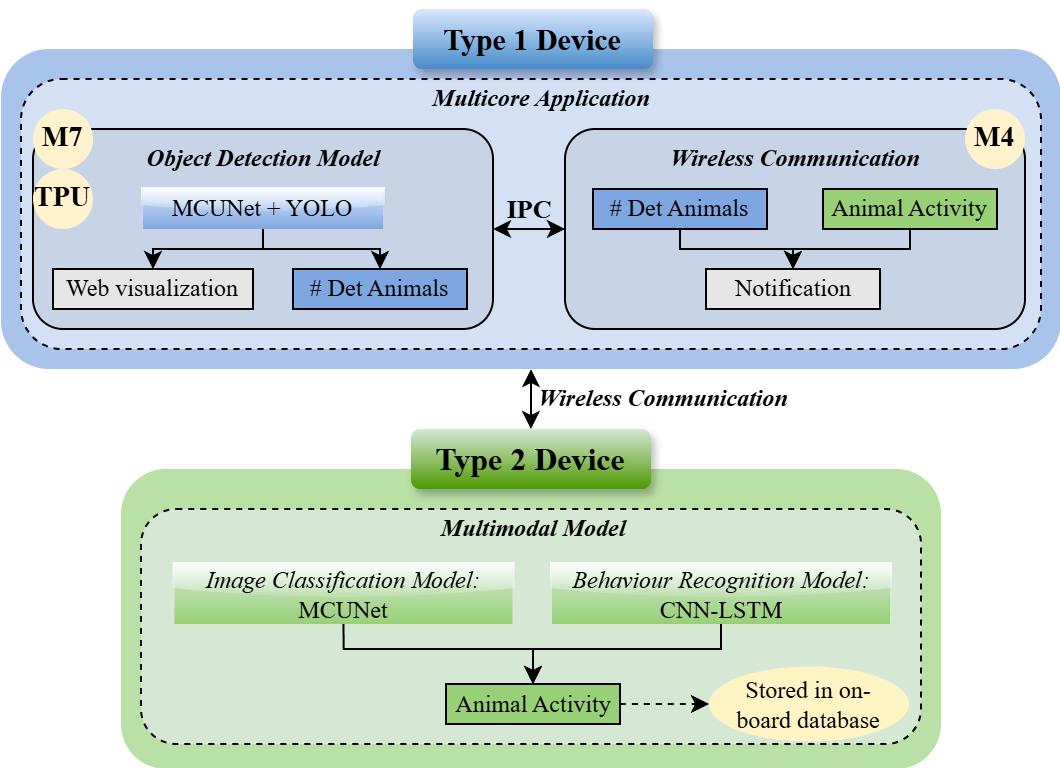}
    \caption{Architecture of the constructed system with multicore application and multimodal network.}
    \label{fig:system_construction}
\end{figure}

The system is constructed by integrating all components developed in the preceding stages, as outlined in Figure \ref{fig:system_construction}. The implementation of the livestock sensing system mainly centers on developing the two types of devices in Figure \ref{fig:monitoring_sys}. The Type 1 device operates a multicore application, which leverages the M7 processor and Edge TPU to perform object detection using the MCUNet-YOLO model. Meanwhile, the M4 processor handles incoming messages from the M7 processor and the Type 2 device through IPC and wireless communication, generating notifications to inform farmers about the animals' status. On the other hand, the Type 2 device utilizes a multimodal network to monitor animal behavior with data from both camera and accelerometer. The fused model is then deployed on the MCU using a single-core application. Additionally, the device also contains an on-board database that stores the animal activity history, providing a reference for the movement tracking and behavior analysis.

\vspace{-0.03\linewidth}

\vspace{-0.03\linewidth}
\section{Evaluation}
\subsection{Evaluation Metrics and Requirements}
The assessment of the individual TinyML models and the complete system was carried out separately on the cloud GPUs and MCU platforms, following metrics in Table \ref{tab:eval-metrics}.

% \vspace{-0.01\linewidth}
\begin{table}[!htbp]
\centering
\caption{Evaluation metrics summary.}
\label{tab:eval-metrics}
% \begin{tabular}{|l|l|}
\begin{tabular}{|c|p{6cm}|}
\hline
\multirow{2}{*}{\textbf{Performance}} &
  \textbf{Accuracy} (\textit{classification} model): $Accuracy = \frac{TP+TN}{TP+TN+FP+FN}$ \\ \cline{2-2} 
 &
  \textbf{mAP} (\textit{object detection} model): 
  $mAP = \frac{1}{N} \sum_{i=1}^{N} AP_i$ where $TP$ is determined by the $IoU = \frac{A_{\text{pred}} \cap A_{\text{gt}}}{A_{\text{pred}} \cup A_{\text{gt}}}$ and threshold. \\ \hline
\multirow{3}{*}{\textbf{Memory Profile}} &
  \textbf{\#FLOPs}: Total number of arithmetic (floating point) operations to make a prediction on a single input, a measure of computational complexity. \\ \cline{2-2} 
 &
  \textbf{\#Params}: Total number of trainable parameters, corresponds to the model size stored on the flash memory based on the variable format (32-bit float, 8-bit integer, \textit{etc.}). \\ \cline{2-2} 
  &
  \textbf{Peak Activation Size}: Maximum amount of dynamic memory usage based on the sum of input and output activations for a layer, constrained by the \textbf{SRAM} size in MCUs. \\ \hline
\textbf{Inference Time} &
  \textbf{End-to-end latency} when running one loop of inference, including input data preprocessing stage, model forward pass and output post-processing steps. This is measured on different processors separately with fully-quantized INT8 models. \\ \hline
\end{tabular}
\end{table}

\vspace{-0.03\linewidth}
\subsection{TinyML Model Evaluation}
\subsubsection{Image Classification}
The performance of the image classification models is displayed in Table \ref{tab:img_classify_models_eval}. Despite the minor accuracy drops with full integer quantization (around 0.5\%), it can significantly improve the inference efficiency and save extensive computational resources by reducing the size of a model, contributing to a network that is highly suitable for MCU deployment. The MCUNet models for both datasets share the same runtime time in Table \ref{tab:img-classify-time}, with a very fast inference speed (60ms) when accelerated with the Edge TPU.
Overall, MCUNet demonstrates competitive performance across different datasets with notably lower operations and trainable parameters (4.8$\times$ lower) compared to the MobileNetV2 \cite{sandler2018mobilenetv2} model.

\begin{table}[!htbp]
\centering
\caption{Image classification models performance with input resolution = 176.}
\label{tab:img_classify_models_eval}
\scalebox{0.8}{
\begin{tabular}{l|l|l|lll}
\hline\hline
\multicolumn{1}{c|}{\multirow{2}{*}{\textbf{Model}}} &
  \multicolumn{1}{c|}{\multirow{2}{*}{\textbf{Dataset}}} &
  \multicolumn{1}{c|}{\multirow{2}{*}{\textbf{\begin{tabular}[c]{@{}c@{}} Acc (\%)\\ (fp32/int8)\end{tabular}}}} &
  \multicolumn{1}{c}{\multirow{2}{*}{\textbf{\#FLOPs}}} &
  \multicolumn{1}{c}{\multirow{2}{*}{\textbf{\#Params}}} &
  \multirow{2}{*}{\textbf{Peak Acts}} \\
\multicolumn{1}{c|}{}   & \multicolumn{1}{c|}{} & \multicolumn{1}{c|}{} & \multicolumn{1}{c}{}    & \multicolumn{1}{c}{} &                       \\ \hline
\multirow{3}{*}{MCUNet} & ImageNet(1k)          & 68.2/67.5             & \multirow{3}{*}{81.64M} & 0.7421M              & \multirow{3}{*}{341K} \\
                        & Mini-ImageNet         & 86.8/86.2             &                         & 0.5925M              &                       \\
                        & Custom Dataset        & 97.7/97.6             &                         & 0.5769M              &                       \\ \hline
MobileNetV2 \cite{sandler2018mobilenetv2}            & ImageNet(1k)          & 71.8/71.0             & 195.5M                 & 3.50M                & 743K             \\ \hline\hline
\end{tabular}
}
\end{table}

\begin{table}[!htbp]
\centering
\caption{Image classification models inference time.}
\label{tab:img-classify-time}
\begin{tabular}{c|cll}
\hline\hline
\multirow{2}{*}{\textbf{Model}} & \multicolumn{3}{c}{\textbf{Avg Inference Time}} \\
 & \textbf{M4} & \multicolumn{1}{c}{\textbf{M7}} & \multicolumn{1}{c}{\textbf{TPU}} \\ \hline
\multicolumn{1}{l|}{MCUNet} & \multicolumn{1}{l}{16316ms} & 2645ms & 60ms \\ \hline\hline
\end{tabular}
\end{table}

% \vspace{-0.03\linewidth}
\subsubsection{Object Detection}
For performance with the Pascal VOC dataset, the optimized MCUNet-YOLO model is compared against the previous research work, ResNet-YOLO \cite{8833671} and the Darknet-YOLOv1 \cite{Redmon_2016_CVPR} networks. As illustrated in Table \ref{tab:obj_det_pascal}, Darknet-YOLOv1 achieves the highest mAP of 0.63 but at the expense of substantial computational demands and the largest model size. The other deep network, ResNet-YOLOv1, delivers a similar mAP to MCUNet-YOLO but at a much higher memory and parameter cost. In contrast, MCUNet-YOLO is the most efficient model with minimal FLOPs and the smallest size. Given the Darknet version is nearly 270$\times$ larger than the MCUNet network, a performance loss of 0.14 is generally acceptable.

The MCUNet-YOLO was eventually deployed and evaluated on the Google Coral Dev Board Micro. As shown in Table \ref{tab:obj_det_time}, running with the M4 processor is highly inefficient, resulting in a latency exceeding 20 seconds due to the model's complexity. Therefore, TPU acceleration is recommended for faster inference (77ms). 

\begin{table}[!htbp]
\centering
\caption{Object detection models performance.}
\label{tab:obj_det_pascal}
\scalebox{0.9}{
\begin{tabular}{l|ll|lll}
\hline\hline
\multicolumn{1}{c|}{\textbf{Model}} &
  \multicolumn{1}{c}{\textbf{Res.}} &
  \multicolumn{1}{c|}{\textbf{mAP}} &
  \multicolumn{1}{c}{\textbf{\#FLOPs}} &
  \multicolumn{1}{c}{\textbf{\#Params}} &
  \multicolumn{1}{c}{\textbf{Peak Acts}} \\ \hline
\rowcolor[HTML]{FFFFC7} 
MCUNet-YOLO   & 224 & 0.49 & 152.4M  & 0.73M   & 0.55M \\
ResNet-YOLO \cite{8833671} & 448 & 0.49 & 16.98G  & 26.73M  & 12.80M  \\
Darknet-YOLO \cite{Redmon_2016_CVPR} & 448 & 0.63 & 20.21G  & 194.23M & 2.81M  \\ \hline\hline
\end{tabular}
}
\end{table}

\begin{table}[!htbp]
\centering
\caption{Object detection model inference time.}
\label{tab:obj_det_time}
\begin{tabular}{c|cll}
\hline\hline
\multirow{2}{*}{\textbf{Model}}      & \multicolumn{3}{c}{\textbf{Avg Inference Time}}                                              \\
                                     & \textbf{M4}             & \multicolumn{1}{c}{\textbf{M7}} & \multicolumn{1}{c}{\textbf{TPU}} \\ \hline
\multicolumn{1}{l|}{MCUNet-YOLO} & \multicolumn{1}{l}{20s+} & 3871ms                          & 77ms                            \\ \hline\hline
\end{tabular}
\end{table}

\vspace{-0.06\linewidth}
\subsection{Behavior Classification}
The evaluation of the CNN-LSTM model is presented in Table \ref{tab:acc_classify_performance}. It demonstrates strong efficiency in categorizing animal behavior based on accelerometer inputs, and achieves a very high accuracy over 98\% with a small number of operations and trainable parameters. Compared to other tasks, running this model on the MCU is much faster, making it suitable for resource-constrained environments.

\begin{table}[!htbp]
\centering
\caption{Behavior recognition model performance.}
\label{tab:acc_classify_performance}
\scalebox{0.75}{
\begin{tabular}{l|l|lll|lll}
\hline\hline
\multicolumn{1}{c|}{\multirow{2}{*}{\textbf{Model}}} &
  \multicolumn{1}{c|}{\multirow{2}{*}{\textbf{\begin{tabular}[c]{@{}c@{}} Acc (\%)\\ (fp32/int8)\end{tabular}}}} &
  \multicolumn{1}{c}{\multirow{2}{*}{\textbf{\#FLOPs}}} &
  \multicolumn{1}{c}{\multirow{2}{*}{\textbf{\#Params}}} &
  \multicolumn{1}{c|}{\multirow{2}{*}{\textbf{Peak Acts}}} &
  \multicolumn{3}{c}{\textbf{Avg Inference Time}} \\
\multicolumn{1}{c|}{} &
  \multicolumn{1}{c|}{} &
  \multicolumn{1}{c}{} &
  \multicolumn{1}{c}{} &
  \multicolumn{1}{c|}{} &
  \multicolumn{1}{c}{\textbf{M4}} &
  \multicolumn{1}{c}{\textbf{M7}} &
  \multicolumn{1}{c}{\textbf{TPU}} \\ \hline
CNN-LSTM &
  98.95/98.94 &
  12.52M & 0.1858M & 31.7K & 2500ms & 495ms & 16ms \\ \hline\hline
\end{tabular}
}
\end{table}

\vspace{-0.06\linewidth}
\subsection{System Evaluation}
The system was assessed on the Google Coral Dev Board Micro, which includes two main applications. The first one is the multimodal network that fuses the image classification and behavior recognition model for comprehensive activity monitoring. The other one focuses on the multicore design, which uses M7 to run the object detection model for animal tracking and M4 for handling wireless communication.

\subsubsection{Multimodal System}
The fused network demonstrates a strong ability to capture animal behaviors by leveraging the multimodal inputs from both sensors and cameras, achieving an average accuracy of over 97\% in Table \ref{tab:fused_model_performance}. Since the late fusion technique was employed to combine the two individual networks, the total size of this application is approximately equal to the sum of these smaller models. Moreover, this system provides an additional function to store animal activity history, which can be retrieved through the on-board flash memory. 

\vspace{-0.04\linewidth}
\begin{table}[!htbp]
\centering
\caption{Fused model performance.}
\label{tab:fused_model_performance}
\scalebox{0.75}{
\begin{tabular}{c|c|ccc|cll}
\hline\hline
\multirow{2}{*}{\textbf{Model}} &
  \multirow{2}{*}{\textbf{\begin{tabular}[c]{@{}c@{}}Acc (\%)\\ (fp32/int8)\end{tabular}}} &
  \multirow{2}{*}{\textbf{\#FLOPs}} &
  \multirow{2}{*}{\textbf{\#Params}} &
  \multirow{2}{*}{\textbf{Peak Acts}} &
  \multicolumn{3}{c}{\textbf{Avg Inference Time}} \\
 &
   &
   &
   &
   &
  \textbf{M4} &
  \multicolumn{1}{c}{\textbf{M7}} &
  \multicolumn{1}{c}{\textbf{TPU}} \\ \hline
\multicolumn{1}{l|}{Fused Model} &
  \multicolumn{1}{l|}{97.56/96.87} &
  \multicolumn{1}{l}{107.3M} &
  \multicolumn{1}{l}{0.6825M} &
  \multicolumn{1}{l|}{341K} &
  \multicolumn{1}{l}{19s} &
  2970ms &
  75ms \\ \hline\hline
\end{tabular}
}
\end{table}

% \vspace{-0.03\linewidth}
\subsubsection{Multicore Applications}
The multicore design fully utilizes the processing power of the two available processors by executing algorithms on both. The total inference time increased by around 20ms compared to running a single object detection network, which is expected since both cores are actively involved.

\subsubsection{Comparison with Existing Livestock Monitoring Systems}
To firmly demonstrate the performance of the system, we compared it against several livestock monitoring applications including sensor-based, vision-based and cloud-based solutions. The multimodal system provides distinctive benefits over single-modality approaches prevalent in the existing applications. Wearable sensor systems \cite{MAO2023108043, LISEUNE2021106566} demonstrate lower power consumption and cost but are confined to accelerometer data, achieving around 90\% accuracy versus our 98\% fused model performance. By incorporating visual inputs from the embedded camera, our system addresses a critical limitation in pure sensor-based solutions where alerts cannot be visually verified.

Similarly, compared to fixed vision-based monitoring methods \cite{10929192, 10502951}, our hybrid architecture combining both stationary (Type 1) and mobile (Type 2) devices that can collectively monitor a group of animals within a wider area, overcoming environmental vulnerabilities while reducing execution latency and power requirements. Although cloud applications can generally achieve better accuracy through dedicated computational resources, our TinyML solution maintains comparable performance of 0.49 mAP through model optimization, while saving hardware cost and improving efficiency. In addition, our approach eliminates the dependency on constant Internet connectivity, benefiting rural areas in contrast to cloud-based systems \cite{10498549} that heavily rely on reliable coverage.

This combination of TPU-accelerated multimodal inference with microcontroller efficiency creates new possibilities for intelligent livestock management in resource-constrained environments where traditional solutions prove impractical.

% \vspace{-0.02\linewidth}
\section{Conclusion}
In summary, we propose a novel and efficient solution for livestock tracking and management in the agriculture sector. With the power of TinyML techniques and comprehensive embedded system development, the entire sensing system is deployed on commercial microcontrollers with minimal latency, enabling on-board inference of various multicore and multimodal AI applications to analyze animal behaviors with TPU acceleration. Since rural areas typically suffer from limited Internet coverage, the system integrates wireless communication frameworks to notify farmers of abnormal activities, ensuring reliable operation in the farm and improving management efficiency. 

Although the system demonstrates highly efficient results, some limitations should be acknowledged. First, the system was only evaluated under a controlled environment rather than real farming settings due to financial constraints, its performance could be affected by the sensor placement and unpredictable animal behaviors in practical deployments. Second, the custom dataset and sensor data are limited by the lack of open-source data, which may fail to cover edge cases in real-world circumstances. In the future, larger and more diverse datasets will be incorporated through field deployments across farms of different scales, and longer temporal windows will be explored to optimize animal behavior recognition. These efforts can help validate sensor placement strategies, improve data capture reliability, and refine the system to enhance its robustness. In conclusion, compared with existing algorithms in livestock farming that utilize cloud-based IoT platforms or specialized hardware to run complex deep learning models, this approach offers a cost-effective and reliable system built on commercial microcontrollers and sensors. This design optimizes farming productivity and efficiency by providing a scalable and non-invasive monitoring solution tailored to the constraints of agricultural environments.

\section*{Acknowledgments}
I would like to thanks to my supervisor Prof. Kanjo Eiman, for her insightful suggestions and continuous help throughout every stage of the project at Imperial College London. Her expertise and assistance have been instrumental in the completion of this project. I would also like to express my sincere gratitude to Brad Patrick for his time and technical support.

\bibliographystyle{IEEEtran}
\bibliography{main}

\end{document}